\documentclass[letterpaper, 10 pt, conference]{ieeeconf}  

\IEEEoverridecommandlockouts                              

\usepackage[ruled,vlined,linesnumbered]{algorithm2e}
\usepackage{graphicx}
\usepackage{xcolor}
\usepackage[dvipsnames]{xcolor}
\usepackage{amsmath,amssymb}
\usepackage{centernot}
\usepackage{stackengine}
\usepackage{multirow}
\DeclareMathSizes{10}{10}{4}{4}
\usepackage{subfigure}
\usepackage{lipsum}

\usepackage{makecell}

\makeatletter
\let\NAT@parse\undefined
\makeatother
\usepackage{hyperref}

\hypersetup{%
  colorlinks=true,%
  linkcolor={black},
  citecolor={blue!65!black},
  urlcolor={blue!80!black},
  bookmarksnumbered=true,%
  bookmarksopen=true}

\title{\LARGE \bf
Multi-Cycle Spatio-Temporal Adaptation in Human-Robot Teaming
}

\begin{document}

\author{Alex Cuellar$^{1}$, Michael Hagenow$^{2}$, Julie Shah$^{1}$
\thanks{$^{1}$MIT CSAIL $^{2}$UW Madison Department of Computer Science}%
}

\maketitle
\thispagestyle{empty}
\pagestyle{empty}

\begin{abstract}
Effective human-robot teaming is crucial for the practical deployment of robots in human workspaces. However, optimizing joint human-robot plans remains a challenge due to the difficulty of modeling individualized human capabilities and preferences. While prior research has leveraged the multi-cycle structure of domains like manufacturing to learn an individual's tendencies and adapt plans over repeated interactions, these techniques typically consider task-level and motion-level adaptation in isolation. Task-level methods optimize allocation and scheduling but often ignore spatial interference in close-proximity scenarios; conversely, motion-level methods focus on collision avoidance while ignoring the broader task context. This paper introduces RAPIDDS, a framework that unifies these approaches by modeling an individual's spatial behavior (motion paths) and temporal behavior (time required to complete tasks) over multiple cycles. RAPIDDS then jointly adapts task schedules and steers diffusion models of robot motions to maximize efficiency and minimize proximity accounting for these individualized models. We demonstrate the importance of this dual adaptation through an ablation study in simulation and a physical robot scenario using a 7-DOF robot arm. Finally, we present a user study ($n=32$) showing significant plan improvement compared to non-adaptive systems across both objective metrics, such as efficiency and proximity, and subjective measures, including fluency and user preference. See this paper's companion video at: \texttt{\textcolor{Blue}{https://youtu.be/55Q3lq1fINs}}. 
\end{abstract}

\section{Introduction} \label{sec:Introduction}
Advances in robotics are enabling practical deployments across diverse domains. However, in applications like manufacturing, robots often cannot fully replace human workers, but should rather seamlessly integrate into existing team dynamics. Significant research has explored various teaming techniques to ensure this integration is fluid and effective \cite{natarajan2023human, carroll2019utility}. One important factor influencing team performance is the quality of the robot's human model. For example, Lasota et al., showed that in close-proximity tasks, humans were objectively safer and subjectively more comfortable if the robot possessed prior knowledge of their behavior \cite{lasota2015analyzing}.

While significant research has developed techniques that react to human behavior on the fly \cite{pupa2023human, unhelkar2020semi, zhang2025relevance, fourie2024manifold}, prior knowledge of an individual's tendencies can help refine models of behavior and create plans that best work with and around the human teammate from the start of an interaction. Therefore, others have demonstrated the benefits of learning individualized models of human behavior over multiple task cycles, using the information learned from past experience to inform future plans \cite{fourie2024real, liu2021coordinating}. For example, knowing an individual's strengths and weaknesses (e.g. dexterity when completing an intricate task) a priori can inform high-level task allocation. Similarly, understanding how a person prefers to move through space (e.g. using right vs left hand on a task) allows a robot to adapt its own motions or schedule tasks to avoid their teammate. Enabling robots to learn and adapt to these individualized tendencies can improve human-robot team effectiveness, safety, and subjective experience.

\begin{figure}
    \centering
    \includegraphics[width=\columnwidth]{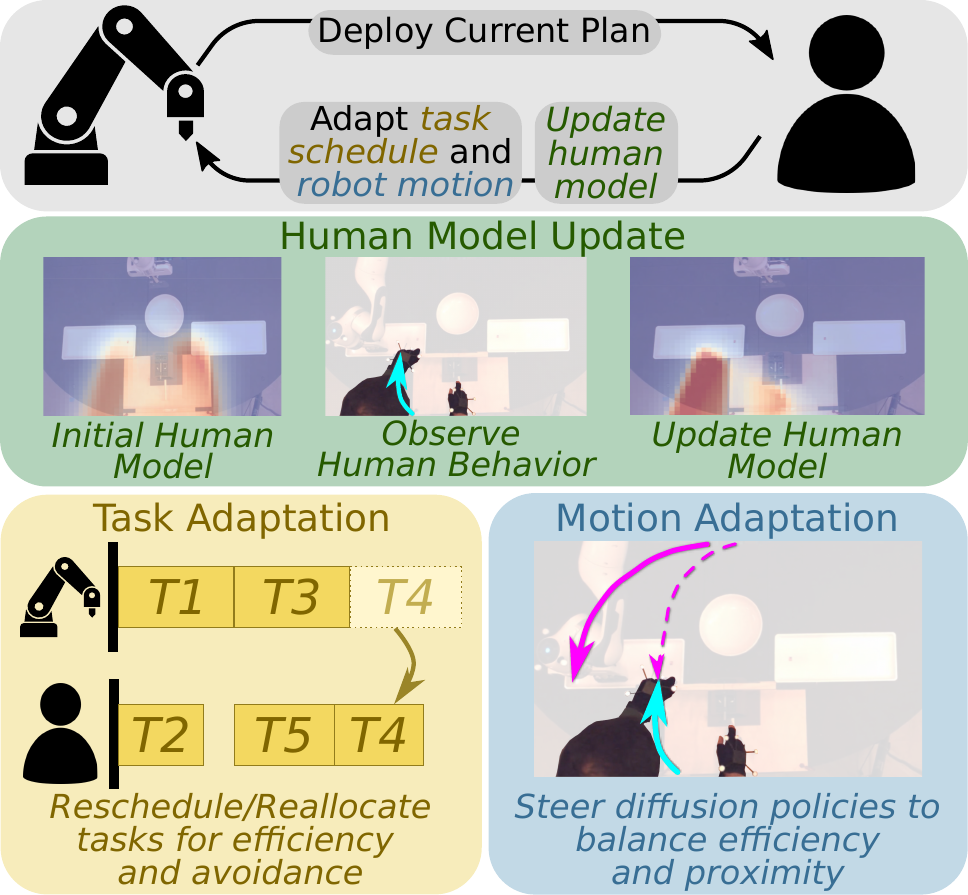}
    \caption{A visualization of the RAPIDDS loop (grey). As the human and robot perform tasks repeatedly, RAPIDDS updates individualized behavior models (green) and jointly adapts schedules (yellow) and motions (blue).}
    \label{fig:Teaser}
\end{figure}

This paper introduces RAPIDDS (\underline{R}epeated \underline{A}daptive \underline{P}lanning via \underline{I}terative \underline{D}eployment of \underline{D}iffusion and \underline{S}cheduling), a framework that (1) learns \textit{individual} models of human movement and duration when completing tasks over repeated iterations and (2) optimizes team plans accounting for this behavior (see Figure \ref{fig:Teaser}). Our adaptation mechanism considers both an individual's task efficiency and their spatial movements. To this end, we formulate a spatial cost function representing the expected proximity between the human and robot and personalize this expected cost via a Bayesian update over successive interactions. The human-robot planner combines task-level schedule optimization with motion-level diffusion steering to maximize efficiency and minimize proximity. Finally, similar to ``exploration-exploitation" tradeoffs in reinforcement learning, early rounds encourage diverse task allocations to learn human behavior on a variety of sub-tasks. This work offers three primary contributions\footnote{See RAPIDDS code at \texttt{\textcolor{Blue}{https://github.com/AlexCuellar/RAPIDDS}}}:

\begin{itemize}
    \item \textbf{A Bayesian adaptation mechanism} that iteratively personalizes models of human spatial and temporal behavior across multiple task cycles.
    \item \textbf{An uncertainty-aware planner} that optimizes joint task schedules and robot motions to maximize efficiency while minimizing expected human-robot proximity.
    \item \textbf{Empirical evidence} that RAPIDDS' spatial and temporal adaptation improves human-robot teaming via a user study considering  efficiency, proximity, and preference.
\end{itemize}

\section{Related Work} \label{sec:Related_Work}
Significant research in human-robot teaming has investigated techniques that adapt to an individual's behavior during interaction.  However, the majority of work considers one task execution in isolation, reasoning over a human's goal \cite{fisac2018probabilistically,zhang2025relevance,cuellar2025alignment}, capability \cite{pupa2022resilient, zhang2020real}, safety \cite{fourie2024manifold, pupa2023human}, or how best to offer assistance \cite{grigore2018preference, vats2025optimal}.  While such techniques demonstrate an ability to react to teammates on the fly, many real world domains involve a predictable repeated structure in which tasks are seen over multilple task cycles (e.g. assembly or food preparation). This structure can be leveraged to learn an individual's behavior and adapt plans more effectively \textit{before} the next cycle.  While understudied, existing work that accounts for this structure (which we refer to as inter-cycle plan adaptation) can be separated into two categories: task-level plan adaptation and motion-level plan adaptation.  

\subsubsection{Task Adaptation} At the task level, inter-cycle adaptation models agents' capabilities, and optimizes task allocation or scheduling to maximize performance metrics (e.g. score in a collaborative game or temporal efficiency). Variants of bandit algorithms, for example, have been proposed to allocate tasks to the highest performing agent in a multi-agent setting while maintaining fairness constraints for team cohesion \cite{chen2020fair, claure2020multi}.  Exploration/exploitation tradeoffs (similar to those inherent to bandit algorithms) have been applied to inter-cycle adaptation focused on schedule optimization as well. Liu et al, for example, learns how long agents in a collaborative setting take to complete sub-tasks toward a wider goal, and use this model to generate more efficient plans in later rounds of interaction \cite{liu2021coordinating}.  Similar to bandit algorithms, an ``entropy" term in the optimization encourages exploration in early interactions, leading to schedules that may not be optimal, but allow the system to learn more about agents' capabilities for better informed schedules in the future.  Despite adaptation at the task level that can increase team effectiveness and efficiency, such methods ignore the spatial aspect of agents acting in close-proximity, possibly leading to spatial interference or over-conservative plans.

\subsubsection{Spatial Adaptation} Motion-level inter-cycle adaptation learns a human teammate's motion while completing a repetitive task, and adapts robot motion to accommodate.  These methods often focus on ``entrainment", a phenomenon in which humans performing a repeated task together often fall into a repeated spatial and temporal pattern \cite{ansermin2017unintentional, fourie2024real}.  Despite roots in cognitive psychology for human-human interaction, studies have shown entrainment extends to human-robot interaction \cite{ansermin2017unintentional}.  Fourie et al utilized the emergent consistency from entrainment to learn a human's preferred path online, and adapt robot motions to best avoid their teammate \cite{fourie2024real}.  However, entrainment is limited to short time-horizon motions repeated precisely, and does not consider task-level adaptation to avoid interference or increase performance.  

RAPIDDS combines the performance benefits from task-level adaptation and avoidance benefits of motion-level adaptation. Via task-level planning, the robot can predict what a human will be doing when, and deploy motions to minimize proximity.  Additionally, via learning an individual's preferred motions, the robot can reschedule tasks and plan motions that optimize for efficiency and minimize spatial interference. 

\subsection{Relationship to Task and Motion Planning}
While this paper focuses primarily on adaptation of team plans, RAPIDDS' task- and motion-level adaptation and planning lends comparison to Task and Motion Planning (TAMP). However, most multi-agent TAMP frameworks generally consider fully controllable agents in largely static environments \cite{shaw2024towards,chen2022cooperative, faroni2023optimal}. In contrast, the uncertainty and uncontrollability of the human teammate requires additional considerations on both the task and motion level \cite{guo2023recent,gottardi2025had}. 

TAMP methods that incorporate a human teammate necessarily relax controllability of the human, and often model uncertainty in aspects of human behavior \cite{akbari2020contingent,gottardi2025had,faroni2020layered}.  However, these approaches limit uncertainty to only the human's task duration, not their motion. Techniques either assume perfect knowledge of human motions a-priori \cite{pellegrinelli2017motion} or plan around the humans in real time \cite{gottardi2025had, akbari2020contingent,zhang2025relevance}.  In contrast, RAPIDDS considers uncertainty in human motion and personalizes a model of spatial separation.  Additionally, to plan tasks requiring complex motion skills, we use a diffusion steering method to find appropriate motions rather than point-to-point trajectories common in TAMP.  
\section{Methods} \label{sec:Methods}
This section details the RAPIDDS framework, which integrates three core components: an individualized model of the human teammate’s spatio-temporal behavior, a task scheduler, and a steerable diffusion motion policy. The human model informs both the scheduler and the diffusion policy. In turn, the scheduler dictates concurrent task allocations to the diffusion model, which then provides the scheduler with robot trajectories and task durations to balance efficiency and proximity. Section \ref{subsec:problem_setting} describes the problem statement. Section \ref{subsec:environment} then describes an example virtual environment used throughout the paper,  followed by Section \ref{subsec:alg_overview} giving an overview of the adaptive planning algorithm central to our contribution. Finally, Section \ref{subsec:scheduling} describes a genetic algorithm to optimize safe and efficient schedules and Section \ref{subsec:Spatial} describes our technique for modeling individual teammates' tendencies when completing tasks. 

\subsection{Problem Setting} \label{subsec:problem_setting}
We model human-robot team planning as a multi-objective optimization problem. The objective function accounts for three factors: total makespan (the time required to complete all tasks), agent distance (a penalty for close human-robot proximity), and diversity. Analogous to ``exploration" in online reinforcement learning, the diversity term encourages varied allocation of tasks to the human during early interaction rounds, enabling the system to learn human tendencies and produce more informed plans in later cycles \cite{liu2021coordinating}.

Let $\boldsymbol{\tau}$ represent the set of tasks to be scheduled, where each task $\tau_i \in \boldsymbol{\tau}$ is defined by its start time $\tau_i^s$ and end time $\tau_i^f$. We define binary assignment variables $A_i^a \in \{0, 1\}$ to indicate whether task $\tau_i$ is assigned to agent $a \in \{h, r\}$. Additionally, integer variables $r_i^a$ track the number of times agent $a$ has completed $\tau_i$ in previous rounds. Precedence constraints are defined by the set $\mathcal{P}$, where a tuple $(i, j) \in \mathcal{P}$ implies that task $\tau_i$ must conclude before $\tau_j$ begins ($\tau_i^f < \tau_j^s$). For human-robot proximity, spatial cost $S_i(\xi^r_j)$ penalizes the expected separation distance between a robot trajectory $\xi^r_j$ and a concurrent human task $\tau_i$. We assume access to a learned diffusion policy $\pi$ that samples robot trajectories:
\begin{align}
    \xi_j^r \sim \pi(j, \{S_i(\cdot) \mid \tau_i \in \mathcal{C}_j\})
\end{align}
Here, each trajectory $\xi$ is a sequence of points $[\mathbf{x}_1, \mathbf{x}_2, \dots]$ where $\mathbf{x}_t \in \mathbb{R}^d$. The diffusion planner is conditioned on task $j$ (via one-hot encoding) and is steered by the spatial costs $\{S_i(\cdot)\}$ of the set of human tasks $\mathcal{C}_j$ expected to occur concurrently with the robot's execution of $\tau_j$ (see Section \ref{subsec:alg_overview}). A complete schedule is defined as the tuple $\mathcal{T} = \langle \boldsymbol{\tau}^r, \boldsymbol{\tau}^h, \boldsymbol{\xi}^r \rangle$, where $\boldsymbol{\tau}^r$ and $\boldsymbol{\tau}^h$ are the task sequences assigned to the robot and human, respectively, and $\boldsymbol{\xi}^r$ contains the corresponding robot trajectories. The optimization is formulated as follows:
\begin{align}
    \underset{\mathcal{T}=\langle \boldsymbol{\tau}^r, \boldsymbol{\tau}^h, \boldsymbol{\xi}^r \rangle}{\min} & \; z_t + \gamma z_s + \lambda z_d \label{eq:total_obj} \\ 
    & z_t = \underset{\tau_i \in \boldsymbol{\tau}}{\max} \; \tau_i^f \label{eq:temporal_obj} \\
    & z_s = \sum_{j : \tau_j \in \boldsymbol{\tau}^r} \underset{i:\tau_i \in \mathcal{C}_j}{\max} S_i(\xi_j^r) \label{eq:space_obj} \\
    z_d = \frac{1}{2|\boldsymbol{\tau}|} \sum_{i=1}^{|\boldsymbol{\tau}|} &\left[ \sum_{a\in \{r,h\}} \left[ \left( \frac{1}{2} \sum_{a'\in \{r,h\}} r_i^{a'} \right) - r_i^{a} \right] \right] \label{eq:div_obj}  \\
    1 = &\sum_{a} A^a_i \label{eq:assign_cond}  \quad \forall \; \tau_i \in \boldsymbol{\tau} \\
    0 = \sum_{i,j,a} A_i^a A_j^a \mathbf{1} &[\max(\tau_i^s,\tau_j^s) < \min(\tau_i^f,\tau_j^f)] \label{eq:no_sim_cond} \\
    &\tau^f_i < \tau^s_j \; \forall \; (i, j) \in \mathcal{P} \label{eq:prec_cond} \\
    \tau^f_i - &\tau^s_i > \,  d_{i}^r \quad \tau_i \in \boldsymbol{\tau}, A^r_i = 1 \label{eq:robot_dur_cond} \\
    \tau^f_i - &\tau^s_i > \,  d_i^h \quad \tau_i \in \boldsymbol{\tau}, A^h_i = 1 \label{eq:human_dur_cond}
\end{align}
The overall objective (Eq \ref{eq:total_obj}) is a linear combination of three objectives (Eqs \ref{eq:temporal_obj} - \ref{eq:div_obj}). The temporal objective (Eq \ref{eq:temporal_obj}) minimizes the total task makespan.  The spatial objective (Eq \ref{eq:space_obj}) minimizes the sum of maximum concurrent spatial cost $S_i(\xi_j^r)$ over robot tasks $\boldsymbol{\tau}^r$. The diversity objective (Eq \ref{eq:div_obj}, called ``entropy" in \cite{liu2021coordinating}) encourages diverse task allocation across multiple cycles. $\gamma$ and $\lambda$ mediate the tradeoff between the three objectives.  To ensure plan validity, the optimization is subject to constraints: Eq \ref{eq:assign_cond} ensures tasks are assigned to exactly one agent; Eq \ref{eq:no_sim_cond} prevents multiple concurrent tasks to one agent; and Eq \ref{eq:prec_cond} enforces precedence orderings.

Equations \ref{eq:robot_dur_cond} and \ref{eq:human_dur_cond} define the task intervals by ensuring the time between start ($\tau_i^s$) and finish ($\tau_i^f$) reflects the respective agent's duration. We treat these duration constraints separately because the scheduler possesses different levels of control and certainty for each agent. Robot durations $d^r_j$ are fixed by the chosen trajectory $\xi_j^r$, whereas human durations are stochastic variables, $d_i^h \sim \mathcal{N}(\mu_i, \sigma^2_i)$, with parameters estimated over cycles of the human-robot task. 
As pointed out by \cite{liu2021coordinating}, this stochasticity means schedule optimization is intractable by MILP methods common in many scheduling domains \cite{gombolay2017computational}. 
Therefore, we take inspiration from \cite{liu2021coordinating}, and instead optimize the task schedule directly over the assignment and ordering of tasks via a genetic algorithm rather than over task start and end times themselves.

\begin{figure}
    \centering
    \includegraphics[width=\columnwidth]{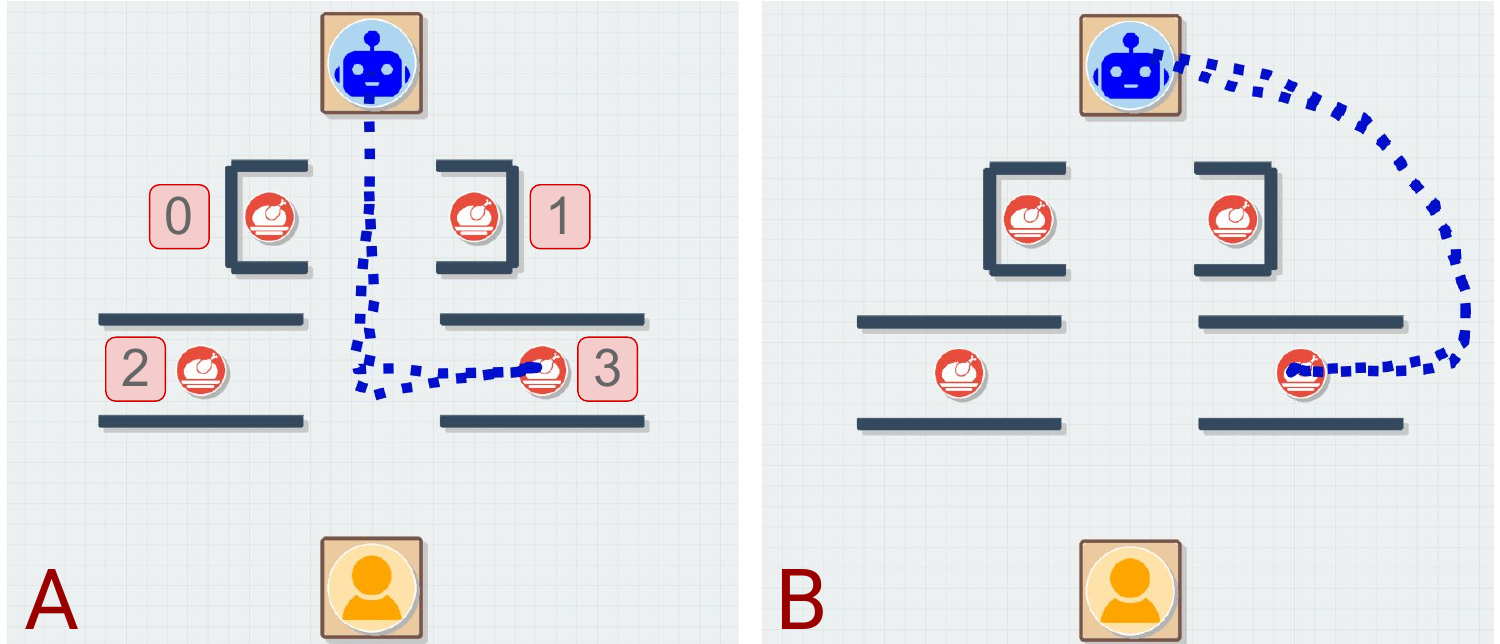}
    \caption{The ``fetch" environment used in this paper.  The human (yellow icon) and robot (blue icon) must collect the objects (red icons) and return them ``home" (brown squares). Figure A shows the task index for fetching each object. Two strategies for $\tau_3$ are shown, \textit{Middle} (A) and \textit{Outside} (B).  }
    \label{fig:Environment}
\end{figure}

\subsection{Example Environment} \label{subsec:environment}
We use a virtual fetching task as a running example (Figure \ref{fig:Environment}). The environment consists of a human player (yellow), a robot player (blue), and four target objects (red) partitioned by walls (black). Agents must fetch objects individually and return them to their respective ``home" locations (brown) before proceeding to the next. The objects are indexed as $\boldsymbol{\tau} = \{\tau_0, \tau_1, \tau_2, \tau_3\}$, with precedence constraints requiring the bottom two objects ($\tau_2, \tau_3$) to be collected before the top two ($\tau_0, \tau_1$). Formally, $\mathcal{P} = \{(\tau_2, \tau_0), (\tau_2, \tau_1), (\tau_3, \tau_0), (\tau_3, \tau_1) \}$. While $\tau_0$ and $\tau_1$ have single-mode traversal paths, $\tau_2$ and $\tau_3$ allow for two distinct motion modes: a ``middle" and an ``outside" strategy (see Figures \ref{fig:Environment}.A and \ref{fig:Environment}.B respectively).
\begin{figure*}
    \centering
    \includegraphics[width=\textwidth]{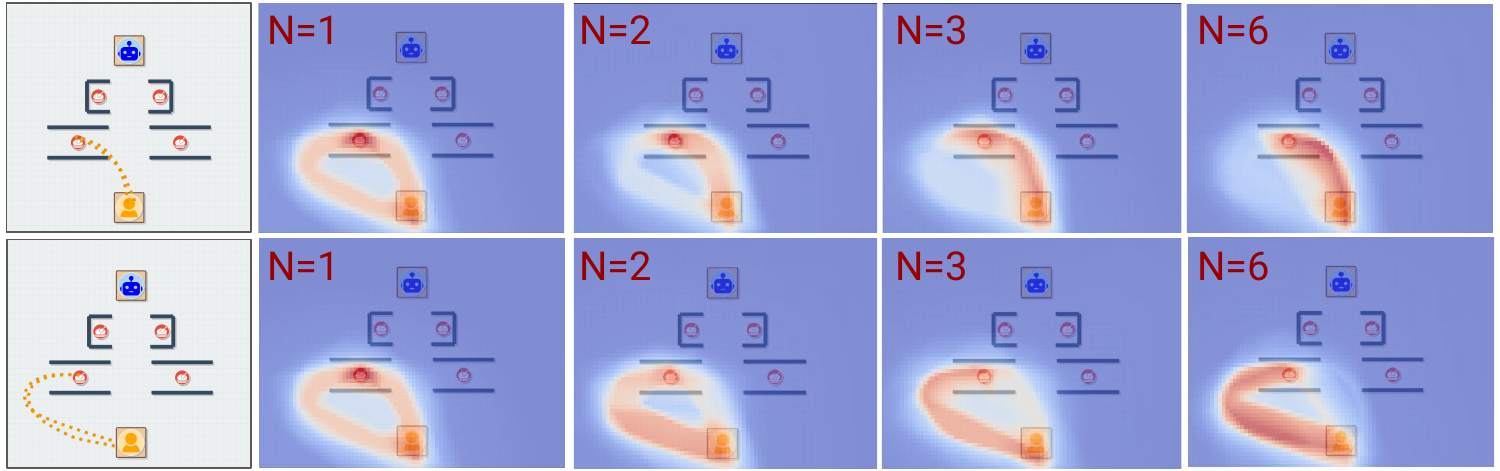}
    \caption{Adaptation of spatial cost $s_i(x)$ over multiple observations of two human strategies for task $\tau_2$: ``middle" (top) and ``outside" (bottom). }
    \label{fig:Spatial_Costs}
\end{figure*}

Even in this simple setup, the human's tendencies (e.g., speed and common path) and the spatial weight $\gamma$ significantly influence the optimal plans. For instance, because $\tau_2$ and $\tau_3$ must be completed first, the robot may initially sample the less efficient but safer ``outside" strategy for $\tau_3$ while the human completes $\tau_2$ to account for uncertainty in the human's motion. Alternatively, a lower $\gamma$ may lead RAPIDDS to prioritize efficiency, sampling the ``middle" strategy despite uncertainty. RAPIDDS balances these trade-offs between proximity and efficiency while iteratively learning an individual's behavior over repeated tasks.

\subsection{Algorithm Overview} \label{subsec:alg_overview}
RAPIDDS aims to learn the distribution over human task durations (i.e. $\boldsymbol{\mu} = \{\mu_1, \mu_2, ..,\}$, $\boldsymbol{\sigma}^2 = \{\sigma^2_1, \sigma^2_2, ... \}$) and spatial tendencies ($S_i(\xi)$) and optimize future schedules to incur the lowest expected cost according to the objective in Eq \ref{eq:total_obj}. Algorithm \ref{alg:overview} describes the adaptation framework.  

The algorithm requires initial human task completion parameters ($\hat{\boldsymbol{\mu}}, \hat{\boldsymbol{\sigma}}^2$), spatial cost functions ($S_i(\xi) \, \forall \, \tau_i \in \boldsymbol{\tau}$), and a diffusion model $\pi$. Each of the $I$ task cycles begins by initializing a memory dictionary $\mathcal{M}$ (Line 3), which maps combinations of concurrent human and robot tasks to specific robot trajectories. Next, a set of $C$ initial candidate schedules is generated using Earliest Deadline First (EDF) scheduling \cite{stankovic1998deadline} (Line 4); while these are guaranteed to be feasible, they are likely sub-optimal. Candidates are then evaluated and ranked via the objective from Eq \ref{eq:total_obj} (line 6). This evaluation involves applying Algorithm \ref{alg:eval}, and determines robot trajectories $\boldsymbol{\xi}^r$ for each candidate. A subset of the highest-performing candidates then undergo an evolution step, which generates a new population of schedules (Line 7; see Section \ref{subsec:scheduling}). This cycle of evaluation and evolution continues for \textit{G} generations, at which point the best plan is deployed (Line 9). During deployment, the human and robot execute their tasks, and the robot records the human's task durations ($d_i^h$) and spatial trajectories ($\xi_i^h$). Finally, the human temporal models ($\hat{\boldsymbol{\mu}}, \hat{\boldsymbol{\sigma}}^2$) and spatial cost functions $S_i(\xi)$ are updated using these observations (Lines 11–12).

\subsection{Genetic Scheduling with Trajectory Adaptation} \label{subsec:scheduling}
Following \cite{liu2021coordinating}, we represent each schedule as a sequence of tasks per agent rather than using precise start and end times. For the environment in Figure \ref{fig:Environment}, a candidate schedule may assign the sequences $\boldsymbol{\tau}^h = [\tau_2, \tau_1]$ and $\boldsymbol{\tau}^r = [\tau_3, \tau_0]$ to the human and robot respectively. In prior work, this representation fully defined a team strategy: because spatial interference is not considered, each agent simply executed their next task as soon as precedence constraints were met. However, our inclusion of a spatial objective renders this representation insufficient. Consider a case where the human’s trajectory for $\tau_1$ overlaps with the robot’s path for $\tau_0$, incurring a high spatial cost. In this scenario, it may be more beneficial for the robot to wait until the human completes $\tau_1$ before beginning $\tau_0$. Standard task sequences cannot explicitly encode this idling. Therefore, we introduce ``wait" tasks $w_i$, which signify that an agent must not proceed to the next step until task $\tau_i$ is finished. In our example, the robot's sequence would be updated to $\boldsymbol{\tau}^r = [\tau_3, w_1, \tau_0]$.  

The evolution step in Algorithm \ref{alg:overview} generates $C$ candidate schedules by ``mutating" the high-performing schedules from the previous optimization round. Liu et al., define three mutation operators—swapping tasks between agents, reordering tasks within an agent’s sequence, and ``crossover" (swapping entire agent sequences after a specific point) \cite{liu2021coordinating}. To manage spatial interference, we introduce two new operators:
\begin{itemize}
    \item \textbf{Add Wait Constraint:} Inserts a wait task $w_i$ at a random position within an agent's sequence.
    \item \textbf{Remove Wait Constraint:} Deletes an existing wait task from a sequence.
\end{itemize}
To ensure schedule validity after mutation, we remove wait tasks if (1) $w_i$ and its target $\tau_i$ are assigned to the same agent, or (2) the wait task is the final step in a schedule.

\begin{algorithm}
\SetAlgoLined
\caption{Multi-Cycle Schedule Adaptation}\label{alg:overview}
\textbf{Input:} \: $\hat{\boldsymbol{\mu}}$, $\hat{\boldsymbol{\sigma}}^2$, $S_i(\xi)$, $\pi$\\
\For{$I$ Cycles}
{
     $\mathcal{M} \leftarrow \{\}$ \\
     $\boldsymbol{\tau}^r,\boldsymbol{\tau}^h$ $\leftarrow$ \{EDF($\hat{\boldsymbol{\mu}}$, $\hat{\boldsymbol{\sigma}}^2$) for i in C\} \\
    \For{$G$ generations}{
        $[z],[\boldsymbol{\xi}^r],\mathcal{M}$ $\leftarrow$ EvalSchedules($[\boldsymbol{\tau}^r],[\boldsymbol{\tau}^h]$,$\hat{\boldsymbol{\mu}}$, $\hat{\boldsymbol{\sigma}}$, $S_i(\cdot)$, $\pi$, $\mathbf{d}^r_{\text{init}}, \mathcal{M}$ ) \\
        $[\mathcal{T}]$ $\leftarrow$ Evolve($[\mathcal{T}]$), $[\mathcal{T}] = [\langle \boldsymbol{\tau}^r, \boldsymbol{\tau}^h, \boldsymbol{\xi}^r \rangle]$
    }
    $\{d_i^h, \xi^h_i \, \forall \, \tau_i \in \boldsymbol{\tau}^h\} \leftarrow$ Deploy($\mathcal{T}_{best}$)  \\
    \For{$\tau_i \in \boldsymbol{\tau}^h$}{
        $\hat{\mu}_i,\hat{\sigma}_i^2 \leftarrow$ Update($d_i^h$) \\
        $S_i(\xi) \leftarrow$ Update($\xi_i^h$)
    }
}
\end{algorithm}

The evaluation step (Algorithm \ref{alg:eval}) assigns an expected cost to each candidate schedule and samples robot trajectories from $\pi$ that optimize the efficiency-proximity tradeoff.  If a schedule violates precedence constraints in $\mathcal{P}$, it is assigned an infinite cost (Line 2). For feasible schedules, the algorithm initializes $\mathcal{C}_j$, the set of human tasks expected to overlap with each robot task $\tau_j$ (Line 3).  Additionally, as sampled trajectories' durations may differ from initial estimates ($\mathbf{d}^r_{\text{init}}$), we initialize an auxiliary overlap set $\mathcal{C}_j'$ to track changes in overlaps (Line 4). Lines 7–13 generate the robot trajectories $\boldsymbol{\xi} = \{\xi_j^r \;\forall \; \tau_j \in \boldsymbol{\tau}^r\}$. To reduce computation, if a trajectory for task $\tau_j$ given the same concurrent tasks $\mathcal{C}_j$ exists in memory $\mathcal{M}$, it is reused (Line 8). Otherwise, we sample a new trajectory from the diffusion policy $\pi$ and store it in $\mathcal{M}$ (Lines 11–12). We use SVDD-PM steering \cite{li2024derivative} to minimize the value function mediating efficiency and proximity:
\begin{align}
V(\xi_j^r) = d^r_j + \gamma \max_{i:\tau_i \in \mathcal{C}_j} S_i(\xi_j^r)
\end{align}
where $d^r_j$ is the duration of sampled trajectory $\xi_j^r$ (Line 15). Since diffusion models sample constant-length sequences, $d^r_j$ is the time at which the task is completed in the trajectory.
\begin{algorithm}
\SetAlgoLined
\caption{Schedule Evaluation}\label{alg:eval}
\textbf{Input:} \: $\boldsymbol{\tau}^r$, $\boldsymbol{\tau}^h$, $\hat{\boldsymbol{\mu}}$, $\hat{\boldsymbol{\sigma}}$, $S_i(\xi)$, $\pi$, $\mathbf{d}^r_{\text{init}}, \mathcal{M}$ \\
 \lIf{Infeasible($\boldsymbol{\tau}^r,\boldsymbol{\tau}^h$)}{\textbf{return} $\infty,$ None, $\mathcal{M}$}
 $\mathcal{C}_j \leftarrow$ Overlaps($\boldsymbol{\tau}^h,\boldsymbol{\tau}^r,\hat{\boldsymbol{\mu}},\mathbf{d}^r_{\text{init}}$) $\forall \; \tau_j \in \boldsymbol{\tau}^r$\\
 $\mathcal{C}_j' \leftarrow  \text{None} \; \forall \; \tau_j \in \boldsymbol{\tau}^r$ \\
\While{$\exists \, \tau_j \in \boldsymbol{\tau}^r \, \text{s.t.} \, \mathcal{C}_j \neq \mathcal{C}_j'$}{
    $\boldsymbol{\xi} \leftarrow \emptyset, \; T \leftarrow \emptyset$ \\
    \For{$\tau_j \in \boldsymbol{\tau}^r$}{
        \uIf{($\tau_j,\mathcal{C}_j) \in \mathcal{M}\text{.keys}$}{
            $\boldsymbol{\xi}[\tau_j] \leftarrow \mathcal{M}[\tau_j,\mathcal{C}_j]$
        }\Else{
            $\boldsymbol{\xi}[\tau_j] \sim \pi(j,\{S_i(\cdot) \; \forall \tau_i \in \mathcal{C}_j\})$ \\
            $\mathcal{M}[\tau_j,\mathcal{C}_j] \leftarrow \boldsymbol{\xi}[\tau_j]$
        }
    }
    $\mathbf{d}^r \leftarrow \text{Durations}(\boldsymbol{\xi})$ \\
    \For{K Iterations}{
        $\mathbf{d}^h \leftarrow \{d_i^h \sim \mathcal{N}(\hat{\mu}_i, \hat{\sigma}^2_i) \; \forall \; \tau_i \in \boldsymbol{\tau}^h\}$ \\
        $T$.add(Makespan($\boldsymbol{\tau}^h,\boldsymbol{\tau}^r, \mathbf{d}^h, \mathbf{d}^r$)) 
    }
    $\mathcal{C}_j' \leftarrow \mathcal{C}_j$ \\
    $\mathcal{C}_j \leftarrow$ Overlaps($\boldsymbol{\tau}^h,\boldsymbol{\tau}^r,\hat{\boldsymbol{\mu}},\mathbf{d}^r$) $\forall \; \tau_j \in \boldsymbol{\tau}^r$ \\
}
$z_t \leftarrow$ Average($T$) \\
$z_s \leftarrow \sum_{j:\tau_j \in \boldsymbol{\tau^r}} \max_{i:\tau_i \in \mathcal{C}_j} S_i(\xi_j^r)$\\
$z = z_t + \gamma z_s + \lambda \text{Diversity}(\boldsymbol{\tau}^h,\boldsymbol{\tau}^r)$ \\
\textbf{return} $z,\boldsymbol{\xi}, \mathcal{M}$
\end{algorithm}

The plan's makespan is then estimated by sampling human task durations $K$ times and averaging the results (Lines 16–19). Based on the updated robot task durations, the overlapping sets $\mathcal{C}_j$ are recalculated (Lines 20–21). If these sets differ from $\mathcal{C}_j'$, trajectory generation repeats with the updated overlaps (Line 5). Once converged, the final temporal, spatial, and diversity costs are calculated and returned alongside the trajectories $\boldsymbol{\xi}$ and updated memory $\mathcal{M}$ (Lines 23–26).

\subsection{Adaptation Mechanism} \label{subsec:Spatial}
\begin{figure*}
      \centering
      \begin{subfigure}
        \centering
        \includegraphics[width = .24\linewidth]{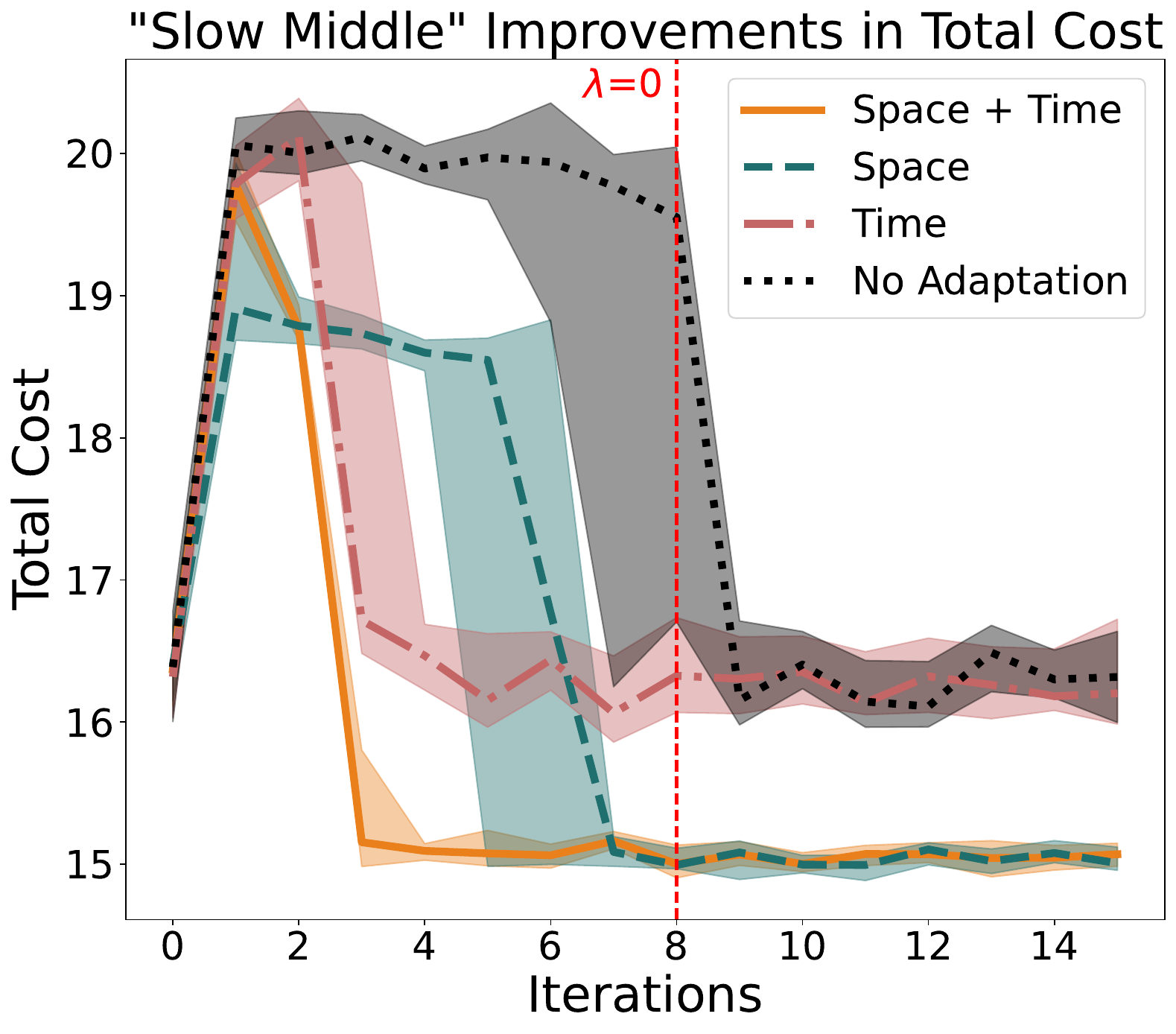}
      \end{subfigure}%
      \hspace{-.2em}
      \begin{subfigure}
        \centering
        \includegraphics[width = .25\linewidth]{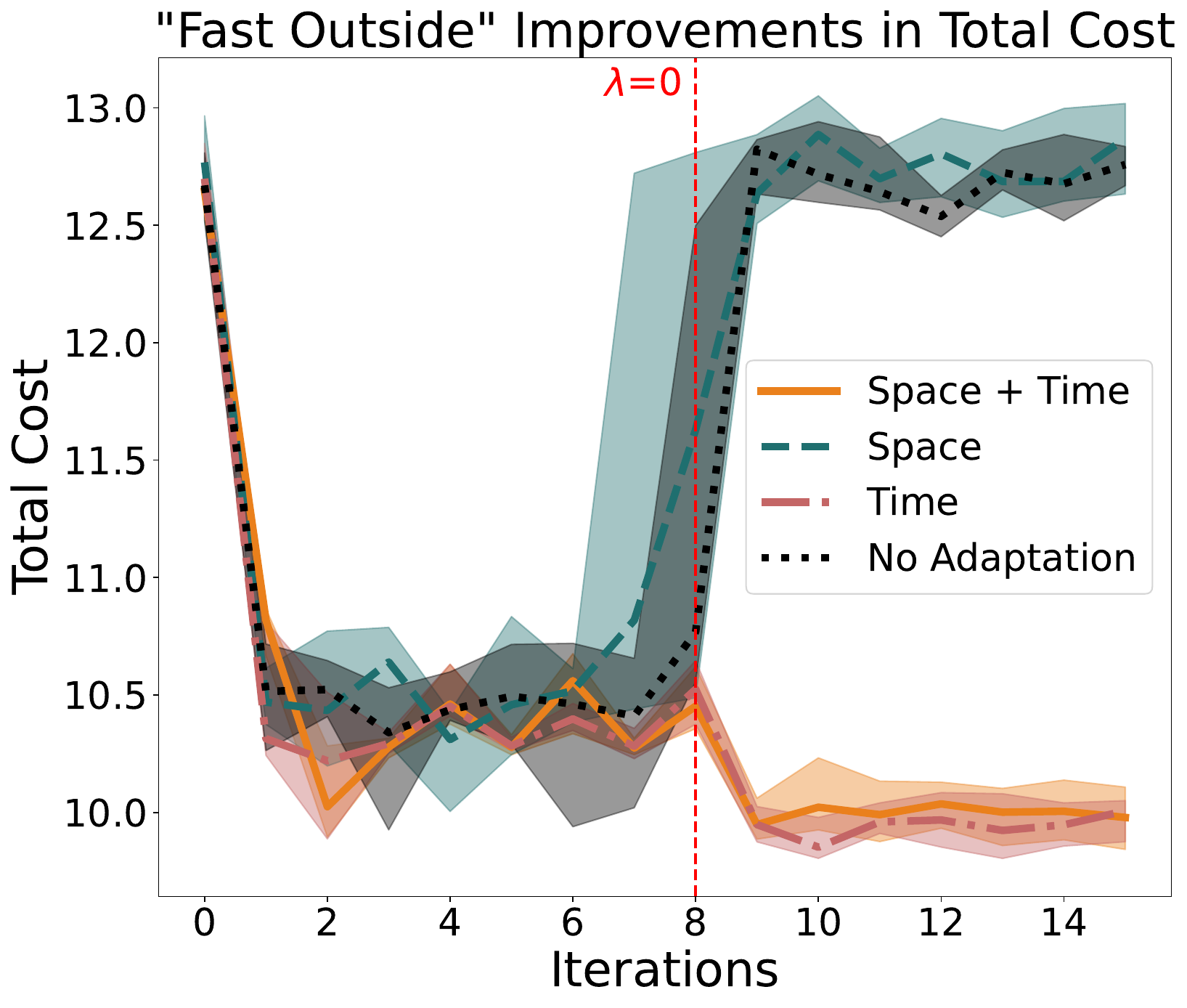}
      \end{subfigure}%
      \hspace{-.2em}
      \begin{subfigure}
        \centering
        \includegraphics[width = .25\linewidth]{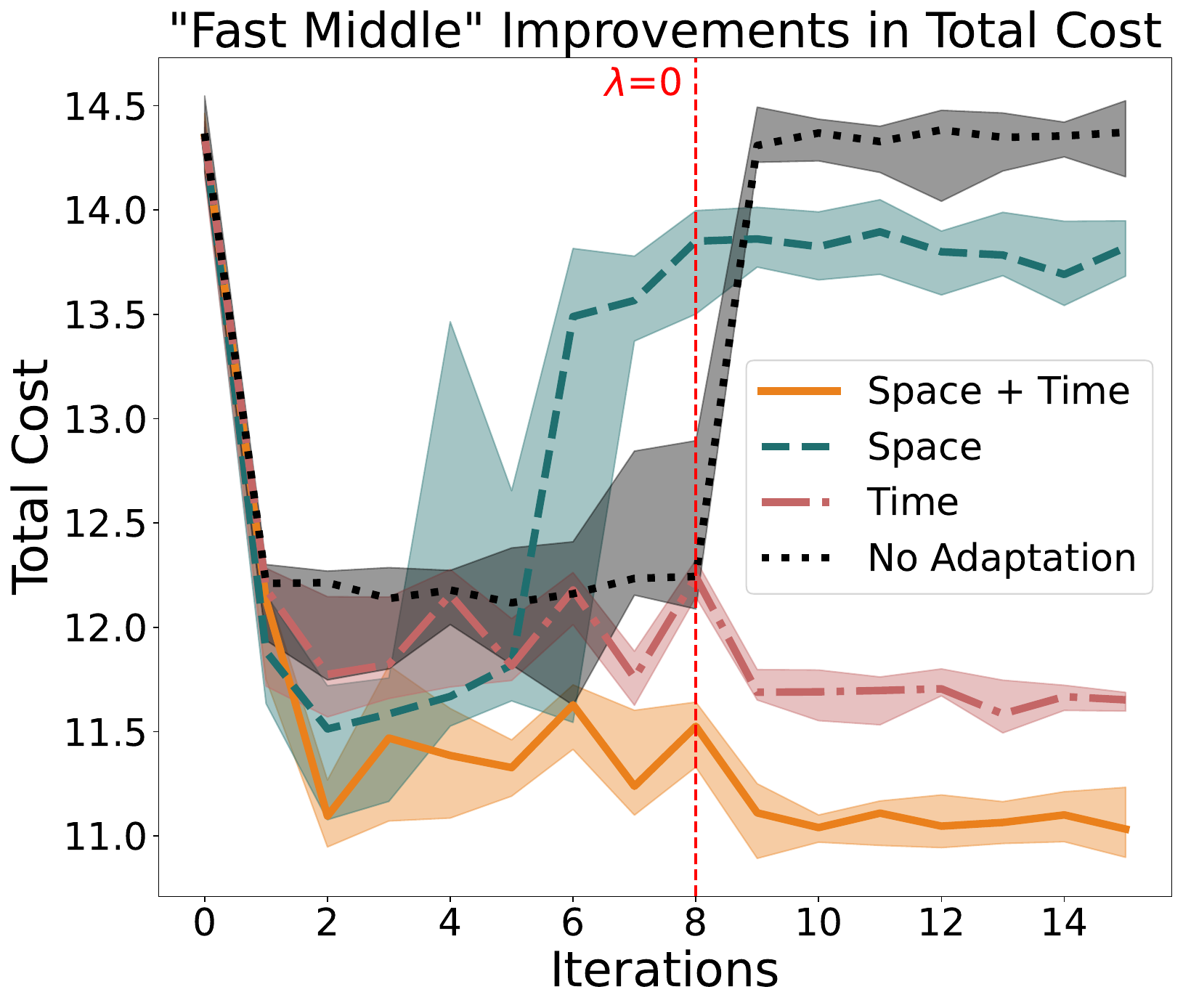}
      \end{subfigure}%
      \hspace{-.2em}
      \begin{subfigure}
        \centering
        \includegraphics[width = .245\linewidth]{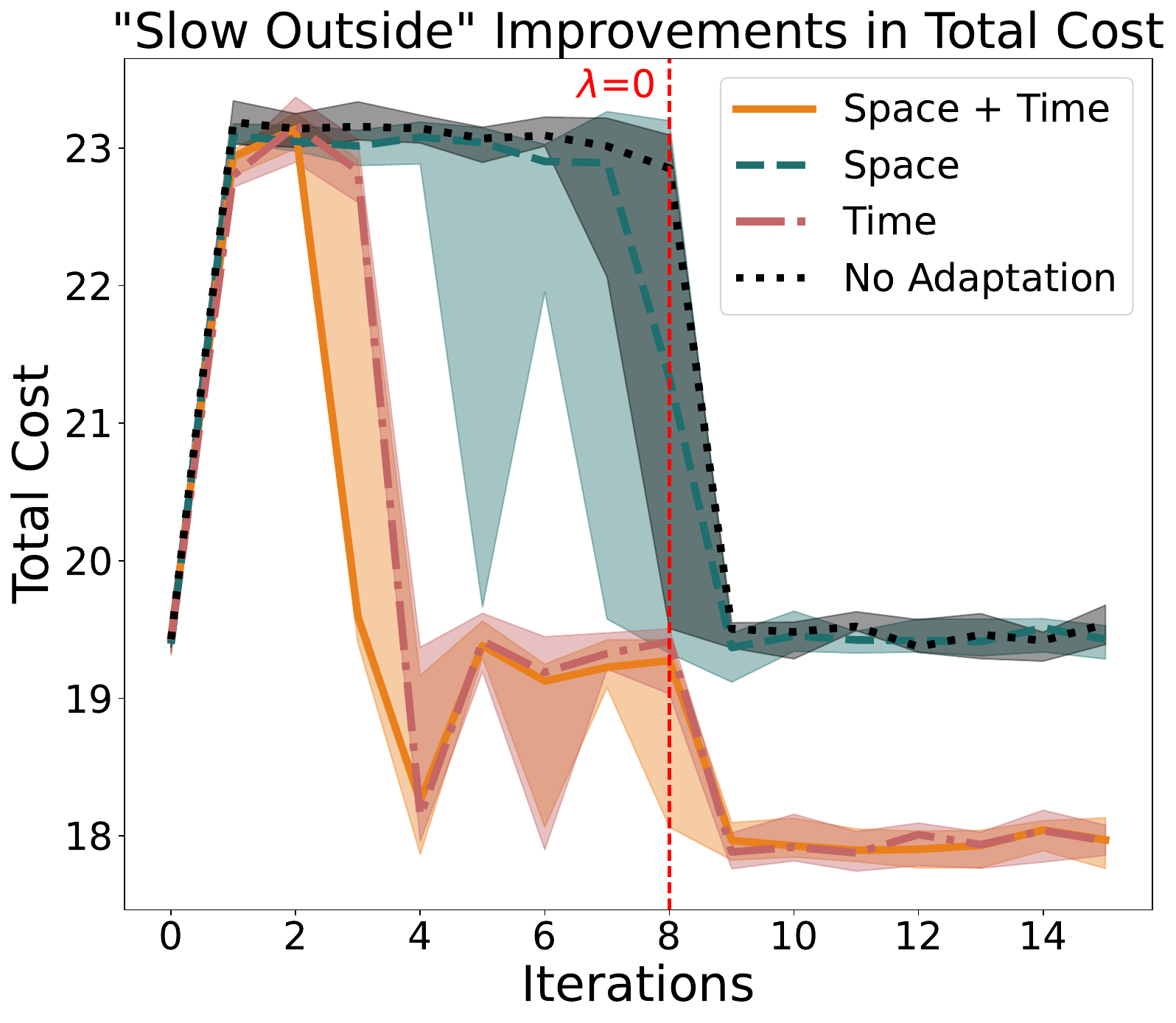}
      \end{subfigure}%
    \caption{Plan cost over 16 cycles of the fetch task (See Section \ref{subsec:environment}) across differing adaptation levels: \textit{space+time} (orange), \textit{space} (blue dashed), \textit{time} (red dot-dash), and \textit{none} (black dotted). We test four human archetypes, left to right: \textit{Fast Outside}, \textit{Slow Middle}, \textit{Fast Middle}, \textit{Slow Outside} }
    \label{fig:AblationResults}
\end{figure*}
In this section, we describe updates to RAPIDDS' model of human behavior spatially (where the teammate is while completing tasks) and temporally (their task durations).
\subsubsection{Spatial Adaptation}
RAPIDDS requires a spatial cost function $S_i(\xi^r)$ to quantify the expected proximity between the human and robot during concurrent robot trajectory $\xi^r$ and human task $\tau_i$. We first define a ``point cost" for a robot position $\mathbf{x}^r \in \mathbb{R}^d$ relative to a human trajectory $\xi^h_i$:
\begin{align} \label{eq:point_cost_fun}
    s_i(\mathbf{x}^r, \xi_h^i) \doteq \max_{\mathbf{x}^h \in \xi^h_i} \exp(-\beta ||\mathbf{x}^h - \mathbf{x}^r||_2).
\end{align}
The expected cost based on the distribution over human trajectories is given by:
\begin{align} \label{eq:expected_point_cost_fun}
    s_i(\mathbf{x}^r) = \mathbb{E}_{\xi_h^i}\left[ s_i(\mathbf{x}^r, \xi^h_i) \right]
\end{align}
This represents the expected negative exponential distance between the point in $\xi^h_i$ closest to robot position $\mathbf{x}^r$. For a new teammate, we initialize the cost function using $M_i$ prior observations from other human teammates ($\xi_i^{p,1} \dots \xi_i^{p,M_i}$) as:
\begin{align} \label{eq:point_cost_fun_init}
    s_i(\mathbf{x}^r) \doteq \frac{1}{M_i}\sum_{m=1}^{M_i} \max_{\mathbf{x}^h \in \xi_i^{p,m}} \exp(-\beta ||\mathbf{x}^h - \mathbf{x}^r||_2)
\end{align}
When observing a new teammate's trajectory $\xi^h_i$ and its resulting cost $s_i(\mathbf{x}^r, \xi_i^h)$, the robot performs a Bayesian update to estimate an individual's ``true cost" $s^*_i(\mathbf{x}^r)$:
\begin{align} \label{eq:point_cost_update}
    p(s^*_i(\mathbf{x}^r) | s_i(\mathbf{x}^r, \xi_i^h)) &= p(s_i(\mathbf{x}^r, \xi_i^h) | s_i^*(\mathbf{x}^r)) p(s_i^*(\mathbf{x}^r)) 
\end{align}
Where the likelihood of an observed trajectory cost is:
\begin{align}
    p(s_i&(\mathbf{x}^r, \xi_i^h) | s_i^*(\mathbf{x}^r)) \sim \mathcal{N}(s^*(\mathbf{x}^r), \rho)
\end{align}
and $\rho$ is a hyperparameter modeling the expected variance of spatial cost for one individual. Figure \ref{fig:Spatial_Costs} shows adaptation of $s_2(\mathbf{x}^r)$ for the fetching task from \ref{subsec:environment} for two individuals.

For computational efficiency, we do not store and determine costs via the full observation history. Instead, we evaluate $s_i(\mathbf{x}^r)$ at $V$ uniformly sampled points across the workspace and approximate the continuous cost function via a neural network. After each cycle, the costs at these $V$ points are updated via Eq. \ref{eq:point_cost_update}, and the networks for all human tasks $\boldsymbol{\tau}^h$ are retrained. In practice, this convergence takes only a few seconds. The trajectory-level cost $S_i(\xi^r)$ is then defined as the maximum point cost along the robot path:
\begin{align}
    S_i(\xi^r) = \max_{\mathbf{x}^r \in \xi^r} s_i(\mathbf{x}^r)
\end{align}
Conceptually, this approximates the negative exponential of the expected minimum distance between any two points in the robot and human trajectories.
\subsubsection{Temporal Adaptation}
Updates to the human temporal parameters $(\hat{\mu}, \hat{\sigma}^2)$ are performed using a joint Bayesian framework. For each task $i$, the posterior distribution of the mean and variance given an observed human duration $d_i^h$ is:
\begin{align}
P(\hat{\mu}_i, \hat{\sigma}_i^2 \mid d_i^h) = \mathcal{N}(d_i^h \mid \hat{\mu}_i, \hat{\sigma}^2_i) P(\hat{\mu}_i, \hat{\sigma}_i^2)
\end{align}
Where the joint prior $P(\hat{\mu}_i, \hat{\sigma}_i^2)$ is a Normal-Inverse-Gamma distribution ($\mu$ is modeled a gaussian and $\sigma$ as an Inverse-Gamma, as is common). Here, $\nu$ serves as a precision parameter informing the model's initial mean $\hat{\mu}_i$, while $\alpha$ and $\beta$ govern the shape and scale of the Inverse-Gamma distribution over the variance $\hat{\sigma}_i^2$. These hyperparameters can be pre-defined or initialized using population-level statistics.




\section{Experiments} \label{sec:Experiments}
In this section, we demonstrate RAPIDDS' adaptation to individuals in teamwork scenarios and the benefits of personalization. Evaluations include an ablation study using the example environment from Section \ref{subsec:environment}, a scenario with a real robot performing a collaborative object painting scenario, and a user study with the same painting task.  

\subsection{Fetch Environment Ablation Study} \label{subsec:Ablation}
We evaluate the RAPIDDS framework via an ablation study using the virtual ``fetch" game (Section \ref{subsec:environment}). We set the initial diversity weight to $\lambda=15$ and spatial weight to $\gamma=1.75$ (which initially encourages the robot to take riskier \textit{Middle} motion; see Figure \ref{fig:Environment}). Following \cite{liu2021coordinating}, $\lambda$ remains constant for the first half of the cycles to encourage diverse task assignment before dropping to zero to prioritize optimal planning. The ablation removes either spatial adaptation (updating only temporal models), temporal adaptation (updating only spatial models), or both. We simulate four human archetypes defined by their movement speed (``fast" vs. ``slow") and spatial strategy for objects 2 and 3 (``middle" vs. ``outside"). ``Middle" and ``outside" motions are shown in the top and bottom rows of Figure \ref{fig:Spatial_Costs} respectively. 

Figure \ref{fig:AblationResults} presents the median and interquartile range (IQR) performance for all adaptation levels and archetypes over 20 trials using 16 interaction cycles per trial. RAPIDDS consistently meets or exceeds the performance of less adaptive systems; specifically, it identifies lower-cost plans than spatial-only adaptation for three archetypes and temporal-only adaptation for two archetypes. In early rounds (before the vertical red line), systems exhibit non-optimal behavior across all scenarios due to the diversity objective. Qualitatively, the improvement of team strategies across archetypes underscores RAPIDDS' ability to generate diverse, personalized plans tailored to specific human behavior.

\begin{figure}
    \centering
    \includegraphics[width=\columnwidth]{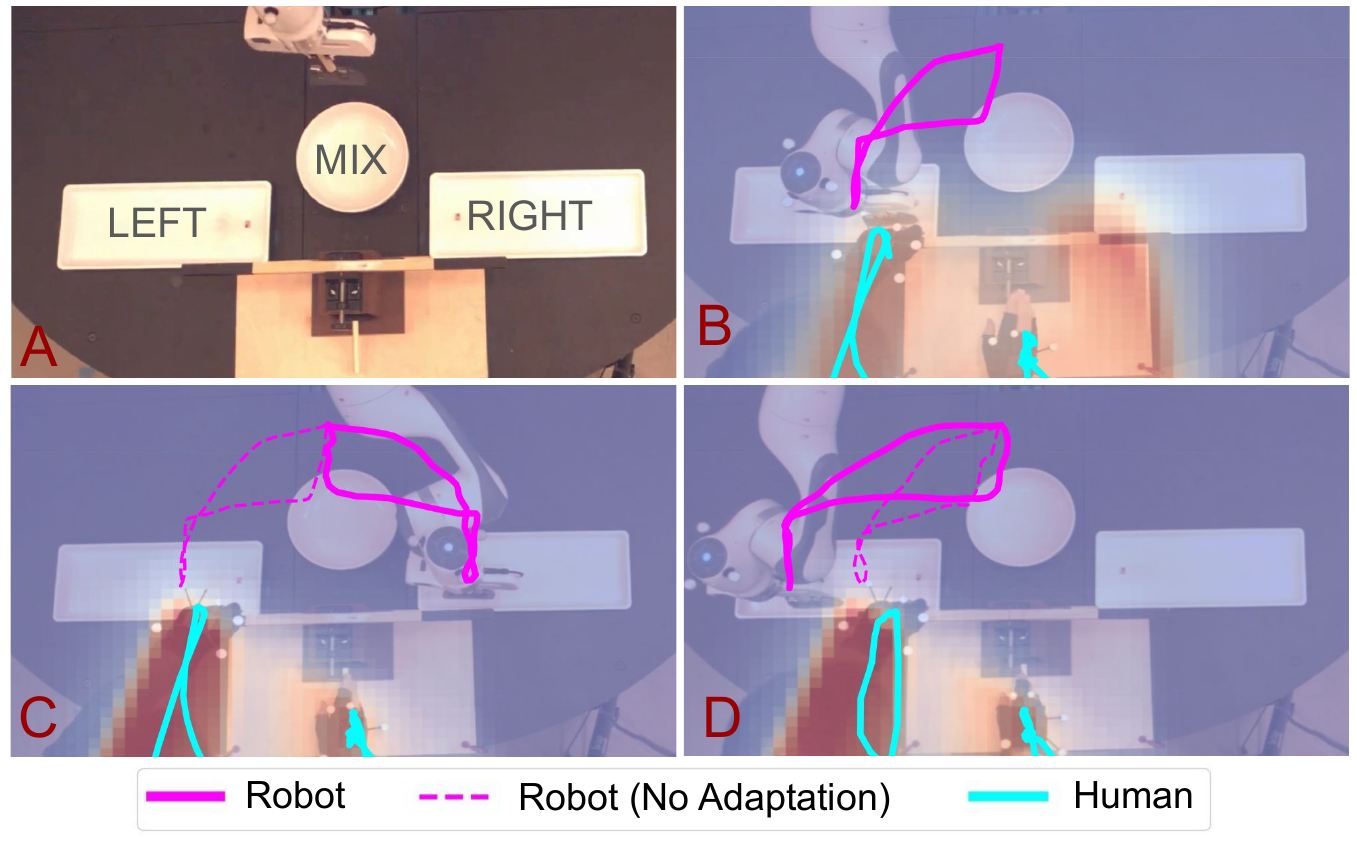}
    \caption{Brush task and robot behavior with RAPIDDS: A) The brush task setting, B) Spatial cost function without adaptation and resulting interference between the human and robot, C) Spatial cost function adapted to left handed user. The \textit{Unconstrained} variant results in switching the order of \textit{Mix L} and \textit{Mix R} to avoid the user. D) Spatial cost function adapted to left handed user. The \textit{Constrained} variant results in less efficient but more avoidant motions.}
    \label{fig:Brush_Setup}
\end{figure}

\subsection{Collaborative Painting} \label{subsec:Case_Study}
To demonstrate RAPIDDS in a physical setting, we designed a human-robot scenario for mixing and applying paint to a workpiece. The process involves five tasks (Figure \ref{fig:Brush_Setup}.A shows the setting, see video for full execution). Two tasks are always assigned to the human: unscrewing a tabletop vice (\textit{Prep Vise}) and securing the workpiece (\textit{Secure Piece}). Two tasks are always assigned to the robot: transferring paint from the left and right trays into a central mixing bowl (\textit{Mix L} and \textit{Mix R} respectively). We investigate two procedural variants: \textit{Unconstrained}, where the robot mixing order is flexible, and \textit{Constrained}, where \textit{Mix R} must precede \textit{Mix L}. Once paint is mixed and the piece is secured, the final task---applying the paint (\textit{Brush})---may be allocated to either teammate.

In this scenario, we explore two kinds of variations in human behavior.  First, individuals vary in hand preference when executing \textit{Secure Piece}. Identifying whether a person uses their left or right hand is critical; for instance, a robot executing \textit{Mix L} may obstruct or come uncomfortably close to a human using their left hand (Figure \ref{fig:Brush_Setup}.B). Because \textit{Secure Piece} follows \textit{Prep Vise}, the robot must commit to a mixing order before observing the human's hand preference within a single cycle. RAPIDDS addresses this by learning an individual’s tendencies over time to optimize planning from the start of each cycle. Second, we vary human brushing rates---which may be slower or faster than the robot---influencing optimal assignment of the \textit{Brush} task to maximize team efficiency.

For simplicity, this discussion assumes the human prefers their left hand for \textit{Secure Piece}. Additionally, we focus discussion of adaptation on learning spatial tendencies, as temporal adaptation has been more extensively studied \cite{liu2021coordinating,gottardi2025had}. We explore RAPIDDS when varying (1) \textit{Constrained} vs \textit{Unconstrained} mix order and (2) spatial weight $\gamma$.  

\begin{figure}
      \centering
      \begin{subfigure}
        \centering
        \includegraphics[width = .5\linewidth]{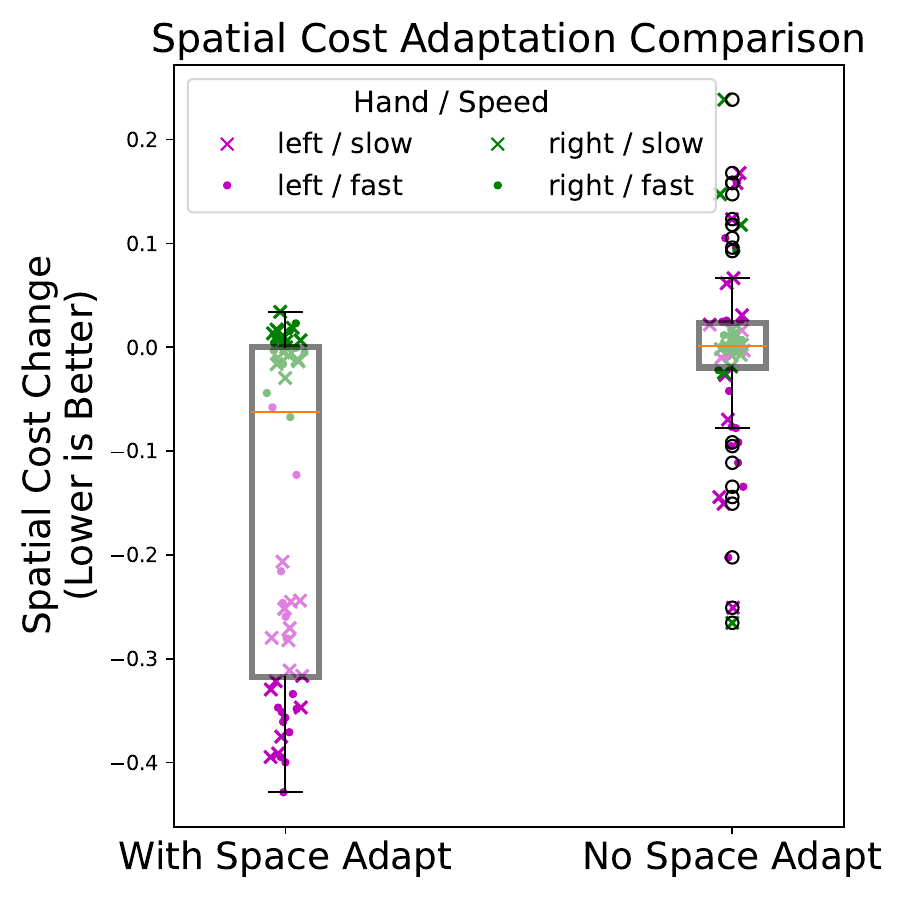}
      \end{subfigure}%
      \hspace{-.6em}
      \begin{subfigure}
        \centering
        \includegraphics[width = .5\linewidth]{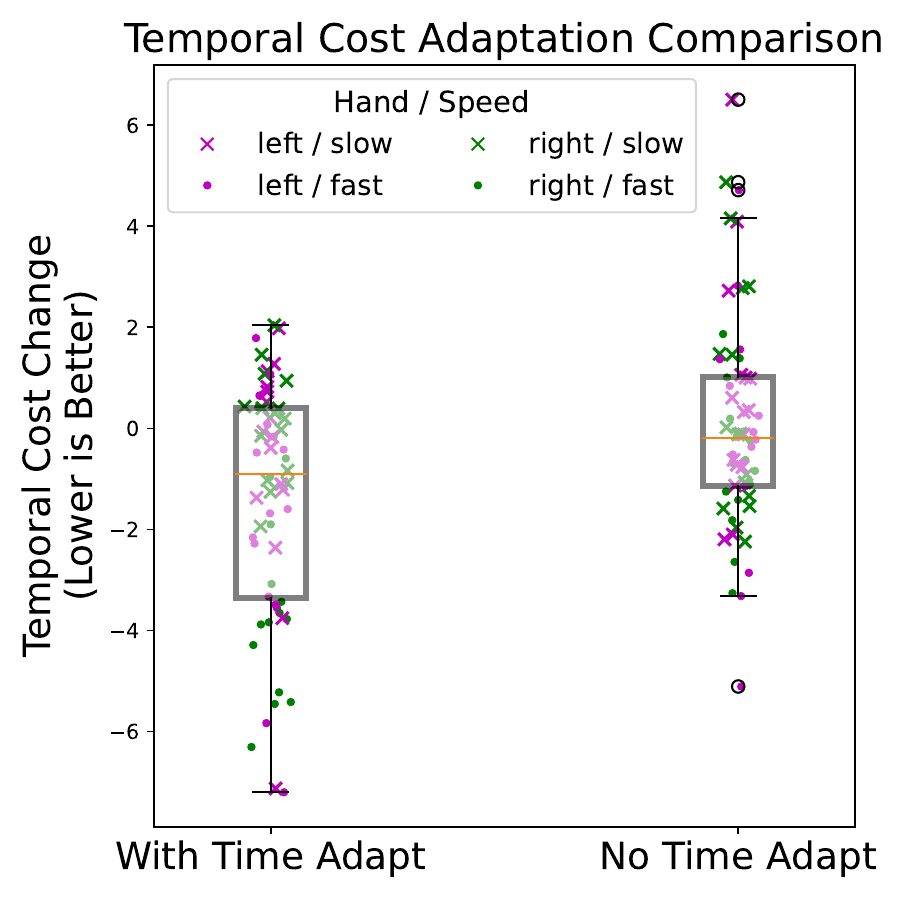}
      \end{subfigure}%
    \caption{User study plan costs with and without adaptation. We separate the effect of spatial adaptation and cost (left) from temporal adaptation and cost (right). Individuals' costs are shown, marked with differences in participant training for speed (magenta vs. green) and donimant hand (``x" vs. dot).}
    \label{fig:Quant_Results}
\end{figure}
\emph{\textbf{Unconstrained, $\gamma=12$:}} Initially, RAPIDDS chooses a random order of \textit{Mix L} and \textit{Mix R} due to uncertainty in the human's preferred hand.  A low $\gamma$ prioritizes efficiency over avoidance, potentially leading to interference (Figure \ref{fig:Brush_Setup}.B).  However, after learning this individual uses their left hand, RAPIDDS schedules \textit{Mix L} first during \textit{Prep Vise}, and thus avoids the person during \textit{Secure Piece} (Figure \ref{fig:Brush_Setup}.C). 

\emph{\textbf{Constrained, $\gamma=12$:}} The first round plays out identically to the prior scenario.  However, since the constrained case must complete \textit{Mix L} after \textit{Mix R} and cannot reorder to avoid the human during \textit{Secure Piece}, diffusion steering samples a less efficient but avoidant motion for \textit{Mix L} (Figure \ref{fig:Brush_Setup}.D). 

\emph{\textbf{Unconstrained, $\gamma=25$:}} With a moderate $\gamma$, the RAPIDDS diffusion steering will choose more conservative but less efficient motions concurrent with \textit{Secure Piece} in the first round due to uncertainty in the human's preferred hand (Figure \ref{fig:Brush_Setup}.D). After learning the human's preference, the planner achieves both low proximity and high efficiency by ordering \textit{Mix L} to occur before \textit{Secure Piece} (Figure \ref{fig:Brush_Setup}.C).

\emph{\textbf{Unconstrained, $\gamma=85$:}} At a high $\gamma$, even conservative motions are deemed too risky under uncertainty. The scheduler initially inserts wait tasks to prevent any robot motion during \textit{Secure Piece}. Once the preference is learned, RAPIDDS resumes the optimal task ordering (\textit{Mix L} then \textit{Mix R}) to ensure safety without idling (Figure \ref{fig:Brush_Setup}.C). 

Even in this relatively simple painting scenario, RAPIDDS can adapt to various human behavior in diverse ways.
\subsection{Collaborative Painting: User Study} \label{subsec:User_Study}
\begin{figure}
    \centering
    \includegraphics[width=\columnwidth]{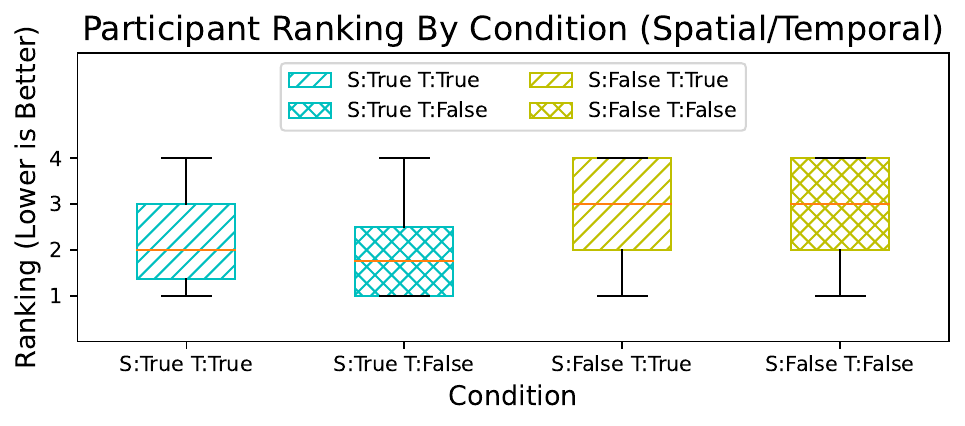}
    \caption{User study participants' rankings of the four systems with differing adaptation levels. Spatially adaptive systems are preferred.}
    \label{fig:Ranking_Results}
\end{figure}
To validate the efficacy of RAPIDDS over multiple interaction cycles, we conducted an IRB-approved user study ($n=32$) using the painting task described in Section \ref{subsec:Case_Study}. After a consent process, participants (aged 18–35, median 22) were trained on the task until achieving consistent performance (typically 5–8 rounds). To simulate diverse real-world behaviors, participants were ``primed" during training: half were instructed to use their right hand for \textit{Secure Piece} while the other half used their left. Additionally, half were trained to perform \textit{Brush} quickly, while the rest performed it slowly. Following training, each participant completed seven rounds of the task with four distinct systems: full adaptation (RAPIDDS), spatial-only adaptation, temporal-only adaptation (equivalent to \cite{liu2021coordinating}), and non-adaptive. System ordering was counterbalanced using a Latin square design. After each system trial, participants completed a fluency survey, which included an item added for physical interference: ``The robot got in the way of me completing my tasks." The study concluded with a forced-choice ranking of the four systems by preference and a demographic survey.

Figure \ref{fig:Quant_Results} illustrates the reduction in spatial and temporal costs, measured as the difference between the first round and the mean of the final three rounds. A two-way repeated measures ANOVA confirmed that spatial adaptation led to a significant decrease in spatial cost ($p < .001$), while temporal adaptation significantly reduced temporal cost ($p < .005$). These results underscore the framework's ability to refine its human model and enable efficient and avoidant interactions.

Figure \ref{fig:Ranking_Results} shows participants' ranking of systems. We see a significantly improved ranking of systems with spatial adaptation (p$<$.005) via a Wilcoxon rank sum test.  Analysis of the survey via an ART ANOVA additionally shows the subjective impact of spatial adaptation, with statements including \textit{``The human-robot team worked fluently together"}, \textit{``The human-robot team improved over time"}, \textit{``The robot's performance improved over time"}, and \textit{``The robot got in the way of me completing my tasks"} showing significant (p$<$0.05) improvement with spatial adaptation.  Such improvements show the importance of spatial adaptation for subjective preference and comfort beyond objective measures.  
\section{Limitations} \label{sec:Limitations}
While RAPIDDS demonstrates clear advantages over systems that adapt to only spatial or temporal patterns, several limitations remain. First, RAPIDDS assumes consistent behavior from a human over time.  However, effects like fatigue and reactions to robot activity may cause shifts in a human's behavior.  Future work may consider methods that recognize and query a teammate concerning shifts in behavior to better model and adapt to future patterns.  

Second, RAPIDDS assumes a static efficiency-proximity tradeoff $\gamma$. However, some individuals in the user study seemed more or less comfortable with close robot proximity.  Therefore, adapating the efficiency-proximity tradeoff itself via communication or observation of the human teammate may be a promising direction for further personalization.  

Finally, while RAPIDDS considers adaptation and planning before each task round, reactivity during task execution remains important and deployed systems would likely benefit from both inter-cycle and reactive on-the-fly adaptation. Future work may consider the integration of reactive planners during task execution with RAPIDDS' inter-cycle planning. 
\section{Conclusion} \label{sec:Conclusion}

This paper introduces RAPIDDS, a framework for multi-cycle spatio-temporal adaption of human-robot team plans.  This framework uses a genetic scheduling technique that optimizes for team efficiency over multiple task cycles while penalizing concurrent tasks that bring the human and robot in close proximity.  Additionally, a diffusion policy selects motions for tasks that are steered to optimize a tradeoff of efficiency and safety.  Finally, we show the benefits of adaptation at both temporal and spatial levels to create task schedules and motions that benefit safety and efficiency.  

\bibliographystyle{IEEEtran}
\bibliography{main.bib}

@article{liu2021coordinating,
  title={Coordinating human-robot teams with dynamic and stochastic task proficiencies},
  author={Liu, Ruisen and Natarajan, Manisha and Gombolay, Matthew C},
  journal={ACM Transactions on Human-Robot Interaction (THRI)},
  volume={11},
  number={1},
  pages={1--42},
  year={2021},
  publisher={ACM New York, NY}
}

@article{lasota2015analyzing,
  title={Analyzing the effects of human-aware motion planning on close-proximity human--robot collaboration},
  author={Lasota, Przemyslaw A and Shah, Julie A},
  journal={Human factors},
  volume={57},
  number={1},
  pages={21--33},
  year={2015},
  publisher={Sage Publications Sage CA: Los Angeles, CA}
}

@inproceedings{unhelkar2020semi,
  title={Semi-supervised learning of decision-making models for human-robot collaboration},
  author={Unhelkar, Vaibhav V and Li, Shen and Shah, Julie A},
  booktitle={conference on Robot Learning},
  pages={192--203},
  year={2020},
  organization={PMLR}
}

@article{pupa2022resilient,
  title={A resilient and effective task scheduling approach for industrial human-robot collaboration},
  author={Pupa, Andrea and Van Dijk, Wietse and Brekelmans, Christiaan and Secchi, Cristian},
  journal={Sensors},
  volume={22},
  number={13},
  pages={4901},
  year={2022},
  publisher={MDPI}
}

@article{fisac2018probabilistically,
  title={Probabilistically safe robot planning with confidence-based human predictions},
  author={Fisac, Jaime F and Bajcsy, Andrea and Herbert, Sylvia L and Fridovich-Keil, David and Wang, Steven and Tomlin, Claire J and Dragan, Anca D},
  journal={arXiv preprint arXiv:1806.00109},
  year={2018}
}

@article{carroll2019utility,
  title={On the utility of learning about humans for human-ai coordination},
  author={Carroll, Micah and Shah, Rohin and Ho, Mark K and Griffiths, Tom and Seshia, Sanjit and Abbeel, Pieter and Dragan, Anca},
  journal={Advances in neural information processing systems},
  volume={32},
  year={2019}
}

@article{natarajan2023human,
  title={Human-robot teaming: grand challenges},
  author={Natarajan, Manisha and Seraj, Esmaeil and Altundas, Batuhan and Paleja, Rohan and Ye, Sean and Chen, Letian and Jensen, Reed and Chang, Kimberlee Chestnut and Gombolay, Matthew},
  journal={Current Robotics Reports},
  volume={4},
  pages={81--100},
  year={2023},
  publisher={Springer},
  link-citations=yes
}

@article{gombolay2017computational,
  title={Computational design of mixed-initiative human--robot teaming that considers human factors: situational awareness, workload, and workflow preferences},
  author={Gombolay, Matthew and Bair, Anna and Huang, Cindy and Shah, Julie},
  journal={The International journal of robotics research},
  volume={36},
  number={5-7},
  pages={597--617},
  year={2017},
  publisher={SAGE Publications Sage UK: London, England}
}

@article{fourie2024manifold,
  title={On-Manifold Strategies for Reactive Dynamical System Modulation with Non-Convex Obstacles},
  author={Fourie, Christopher K and Figueroa, Nadia and Shah, Julie A},
  journal={IEEE Transactions on Robotics},
  year={2024},
  publisher={IEEE}
}

@incollection{pupa2023human,
  title={A human-centered dynamic task scheduling and safe task execution approach for human-robot collaboration scenarios},
  author={Pupa, Andrea and Arrfou, Mohammad and Andreoni, Gildo and Secchi, Cristian},
  year={2023},
  publisher={IET}
}

@inproceedings{grigore2018preference,
  title={Preference-based assistance prediction for human-robot collaboration tasks},
  author={Grigore, Elena Corina and Roncone, Alessandro and Mangin, Olivier and Scassellati, Brian},
  booktitle={2018 IEEE/RSJ International Conference on Intelligent Robots and Systems (IROS)},
  pages={4441--4448},
  year={2018},
  organization={IEEE}
}

@inproceedings{chen2020fair,
  title={Fair contextual multi-armed bandits: Theory and experiments},
  author={Chen, Yifang and Cuellar, Alex and Luo, Haipeng and Modi, Jignesh and Nemlekar, Heramb and Nikolaidis, Stefanos},
  booktitle={Conference on Uncertainty in Artificial Intelligence},
  pages={181--190},
  year={2020},
  organization={PMLR}
}

@inproceedings{claure2020multi,
  title={Multi-armed bandits with fairness constraints for distributing resources to human teammates},
  author={Claure, Houston and Chen, Yifang and Modi, Jignesh and Jung, Malte and Nikolaidis, Stefanos},
  booktitle={Proceedings of the 2020 ACM/IEEE International Conference on Human-Robot Interaction},
  pages={299--308},
  year={2020}
}

@inproceedings{zhang2020real,
  title={Real-time adaptive assembly scheduling in human-multi-robot collaboration according to human capability},
  author={Zhang, Shaobo and Chen, Yi and Zhang, Jun and Jia, Yunyi},
  booktitle={2020 IEEE International Conference on Robotics and Automation (ICRA)},
  pages={3860--3866},
  year={2020},
  organization={IEEE}
}

@inproceedings{ansermin2017unintentional,
  title={Unintentional entrainment effect in a context of Human Robot Interaction: an experimental study},
  author={Ansermin, Eva and Mostafaoui, Ghiles and Sargentini, Xavier and Gaussier, Philippe},
  booktitle={2017 26th IEEE international symposium on Robot and Human Interactive Communication (RO-MAN)},
  pages={1108--1114},
  year={2017},
  organization={IEEE}
}

@phdthesis{fourie2024real,
  title={Real-Time Anticipation and Entrainment in Human-Robot Interaction},
  author={Fourie, Christopher},
  year={2024},
  school={Massachusetts Institute of Technology}
}

@article{li2024derivative,
  title={Derivative-free guidance in continuous and discrete diffusion models with soft value-based decoding},
  author={Li, Xiner and Zhao, Yulai and Wang, Chenyu and Scalia, Gabriele and Eraslan, Gokcen and Nair, Surag and Biancalani, Tommaso and Ji, Shuiwang and Regev, Aviv and Levine, Sergey and others},
  journal={arXiv preprint arXiv:2408.08252},
  year={2024}
}

@article{gottardi2025had,
  title={HAD-TAMP: Human adaptive task and motion planning for human-robot collaboration in industrial scenario},
  author={Gottardi, Alberto and Terreran, Matteo and Pagello, Enrico and Menegatti, Emanuele},
  journal={Robotics and Autonomous Systems},
  pages={105318},
  year={2025},
  publisher={Elsevier}
}

@article{pellegrinelli2017motion,
  title={Motion planning and scheduling for human and industrial-robot collaboration},
  author={Pellegrinelli, Stefania and Orlandini, Andrea and Pedrocchi, Nicola and Umbrico, Alessandro and Tolio, Tullio},
  journal={CIRP Annals},
  volume={66},
  number={1},
  pages={1--4},
  year={2017},
  publisher={Elsevier}
}

@article{guo2023recent,
  title={Recent trends in task and motion planning for robotics: A survey},
  author={Guo, Huihui and Wu, Fan and Qin, Yunchuan and Li, Ruihui and Li, Keqin and Li, Kenli},
  journal={ACM Computing Surveys},
  volume={55},
  number={13s},
  pages={1--36},
  year={2023},
  publisher={ACM New York, NY}
}

@article{shaw2024towards,
  title={Towards practical finite sample bounds for motion planning in TAMP},
  author={Shaw, Seiji and Curtis, Aidan and Kaelbling, Leslie Pack and Lozano-P{\'e}rez, Tom{\'a}s and Roy, Nicholas},
  journal={arXiv preprint arXiv:2407.17394},
  year={2024}
}

@article{chen2022cooperative,
  title={Cooperative task and motion planning for multi-arm assembly systems},
  author={Chen, Jingkai and Li, Jiaoyang and Huang, Yijiang and Garrett, Caelan and Sun, Dawei and Fan, Chuchu and Hofmann, Andreas and Mueller, Caitlin and Koenig, Sven and Williams, Brian C},
  journal={arXiv preprint arXiv:2203.02475},
  year={2022}
}

@article{faroni2023optimal,
  title={Optimal task and motion planning and execution for multiagent systems in dynamic environments},
  author={Faroni, Marco and Umbrico, Alessandro and Beschi, Manuel and Orlandini, Andrea and Cesta, Amedeo and Pedrocchi, Nicola},
  journal={IEEE Transactions on Cybernetics},
  volume={54},
  number={6},
  pages={3366--3377},
  year={2023},
  publisher={IEEE}
}

@article{akbari2020contingent,
  title={Contingent task and motion planning under uncertainty for human--robot interactions},
  author={Akbari, Aliakbar and Diab, Mohammed and Rosell, Jan},
  journal={Applied Sciences},
  volume={10},
  number={5},
  pages={1665},
  year={2020},
  publisher={MDPI}
}

@inproceedings{faroni2020layered,
  title={A layered control approach to human-aware task and motion planning for human-robot collaboration},
  author={Faroni, Marco and Beschi, Manuel and Ghidini, Stefano and Pedrocchi, Nicola and Umbrico, Alessandro and Orlandini, Andrea and Cesta, Amedeo},
  booktitle={2020 29th IEEE international conference on robot and human interactive communication (RO-MAN)},
  pages={1204--1210},
  year={2020},
  organization={IEEE}
}

@article{cuellar2025alignment,
  title={An Alignment-Based Approach to Learning Motions From Demonstrations},
  author={Cuellar, Alex and Fourie, Christopher K and Shah, Julie A},
  journal={IEEE Robotics and Automation Letters},
  year={2025},
  publisher={IEEE}
}

@article{vats2025optimal,
  title={Optimal Interactive Learning on the Job via Facility Location Planning},
  author={Vats, Shivam and Zhao, Michelle and Callaghan, Patrick and Jia, Mingxi and Likhachev, Maxim and Kroemer, Oliver and Konidaris, George},
  journal={arXiv preprint arXiv:2505.00490},
  year={2025}
}

@inproceedings{zhang2025relevance,
  title={Relevance-driven decision making for safer and more efficient human robot collaboration},
  author={Zhang, Xiaotong and Huang, Dingcheng and Youcef-Toumi, Kamal},
  booktitle={2025 IEEE International Conference on Robotics and Automation (ICRA)},
  pages={5899--5905},
  year={2025},
  organization={IEEE}
}

@book{stankovic1998deadline,
  title={Deadline scheduling for real-time systems: EDF and related algorithms},
  author={Stankovic, John A and Spuri, Marco and Ramamritham, Krithi and Buttazzo, Giorgio},
  volume={460},
  year={1998},
  publisher={Springer Science \& Business Media}
}

\end{document}